\documentclass[conference]{IEEEtran}
\IEEEoverridecommandlockouts
\usepackage{cite}
\usepackage{amsmath,amssymb,amsfonts}
\usepackage{algorithmic}
\usepackage{graphicx}
\usepackage{textcomp}
\usepackage{xcolor}
\usepackage{hyperref}
\usepackage{orcidlink}
\usepackage{booktabs}
\usepackage{comment}
\usepackage{float}
\usepackage{listings}
\usepackage{xcolor}
\usepackage[utf8]{inputenc}

\lstdefinelanguage{JSON}{
    basicstyle=\ttfamily\footnotesize,
    stringstyle=\color{orange},
    commentstyle=\color{gray}\itshape,
    keywordstyle=\color{blue}\bfseries,
    morestring=[b]",
    morecomment=[l]{//},
    morecomment=[s]{/*}{*/},
    morekeywords={true,false,null}
}

\def\BibTeX{{\rm B\kern-.05em{\sc i\kern-.025em b}\kern-.08em
    T\kern-.1667em\lower.7ex\hbox{E}\kern-.125emX}}
\begin{document}

\title{Scaling Vision Transformers: Evaluating DeepSpeed for Image-Centric Workloads\\}

\author{
\IEEEauthorblockN{Huy Trinh \orcidlink{0003-4652-3870}}
\IEEEauthorblockA{\textit{Electrical \& Computer Eng} \\
\textit{University of Waterloo}\\
h3trinh@uwaterloo.ca}
\and
\IEEEauthorblockN{Rebecca Ma}
\IEEEauthorblockA{\textit{Electrical \& Computer Eng} \\
\textit{University of Waterloo}\\
rebecca.ma@uwaterloo.ca}
\and
\IEEEauthorblockN{Zeqi Yu}
\IEEEauthorblockA{\textit{Electrical \& Computer Eng} \\
\textit{University of Waterloo}\\
zeqi.yu@uwaterloo.ca}
\and
\IEEEauthorblockN{Tahsin Reza}
\IEEEauthorblockA{\textit{Electrical \& Computer Eng} \\
\textit{University of Waterloo}\\
tahsin.reza@uwaterloo.ca}
}

\maketitle

\begin{abstract}
Vision Transformers (ViTs) have demonstrated remarkable potential in image processing tasks by utilizing self-attention mechanisms to capture global relationships within data. However, their scalability is hindered by significant computational and memory demands, especially for large-scale models with many parameters. This study aims to leverage DeepSpeed, a highly efficient distributed training framework that is commonly used for language models, to enhance the scalability and performance of ViTs. We evaluate intra- and inter-node training efficiency across multiple GPU configurations on various datasets like CIFAR-10 and CIFAR-100, exploring the impact of distributed data parallelism on training speed, communication overhead, and overall scalability (strong and weak scaling). By systematically varying software parameters, such as batch size and gradient accumulation, we identify key factors influencing performance of distributed training. The experiments in this study provide a foundational basis for applying DeepSpeed to image-related tasks. Future work will extend these investigations to deepen our understanding of DeepSpeed's limitations and explore strategies for optimizing distributed training pipelines for Vision Transformers.
\end{abstract}


\section{Introduction}
\begin{figure}[htbp]
\centerline{\includegraphics[width=\linewidth]{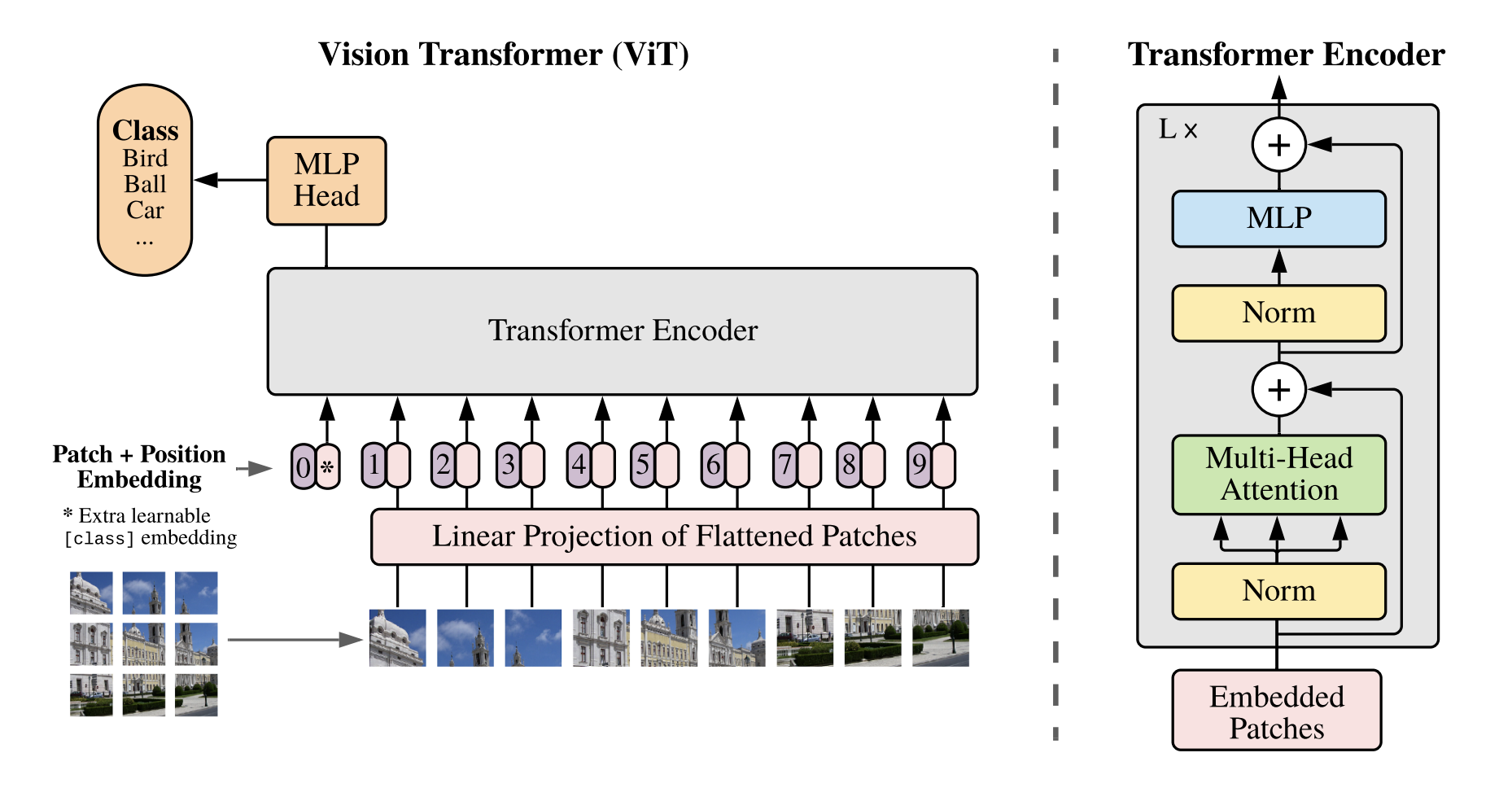}}
\caption{ViT Model Overview \cite{b1}}
\label{ViT}
\end{figure}

Deep Learning (DL) has transformed numerous fields over the past decade, from healthcare to autonomous systems to natural language processing (NLP) and computer vision (CV). Transformers is a deep learning architecture initially introduced for NLP in language models such as BERT and GPT to handle large-scale data, learn complex representations, and generalize across tasks \cite{transformers}. Studies have shown that transformers can also be expanded to apply not only to language tasks but also to image tasks in the form of Vision Transformers (ViT) in applications including image classification and object detection \cite{vision_transformers}. Unlike traditional Convolutional Neural Networks (CNNs) that rely on convolutional layers to extract features hierarchically from the data, transformers adopt the self-attention  mechanism to gain a holistic view of the image. In ViTs shown in Figure~\ref{ViT}, the image is first divided into smaller fixed-size patches and then converted into "tokens" \cite{b1}. These "tokens" are then linearly embedded into a sequence of feature vectors and fed to a standard Transformer encoder \cite{b1}.

The key differentiation between ViTs and traditional CNNs is their ability to model global relationships across an image through the self-attention mechanism. ViTs analyze interactions between all patches of the input data simultaneously, which captures long-range dependencies and provides a holistic understanding of the image. However, a downside of ViTs is the computation cost and memory requirements, especially as the model and data sizes grow (e.g., parameters scaling from millions to billions) \cite{ds_inference}. DeepSpeed is a framework that provides memory-efficient data parallelism, leveraging multiple machines/GPUs to train the model using less time \cite{b3}. Data parallelism replicates the model on each device and performs training on different batches of data in parallel, which is efficient because each machine is training on a portion of the original data. Previous work with DeepSpeed has resulted in impressive results compared to the state-of-art methods, such as training BERT-large in 44 minutes using 1024 V100 GPUs and training GPT-2 with 1.5B parameters 3.75x faster than NVIDIA Megatron on Azure GPUs \cite{b3}. However, DeepSpeed has not been commonly applied to image-related tasks or ViTs. Therefore, this study aims to leverage DeepSpeed and data parallelism to investigate the scalability of ViTs.

Our key contributions are as follows:
\begin{itemize}
    \item Adapt DeepSpeed for Vision Transformers and run inter- and intra-node training to observe scalability trends.
    \item Experiment with distributed training (specifically data parallelism) and measure training speed and communication overhead for increasing number of GPUs. 
    \item Evaluate how changes in software parameters (e.g., batch size, accumulation step) can affect the scalability of the system.
\end{itemize}

\section{Background and Related Work}
\begin{figure}[htbp]
\centerline{\includegraphics[width=\linewidth]{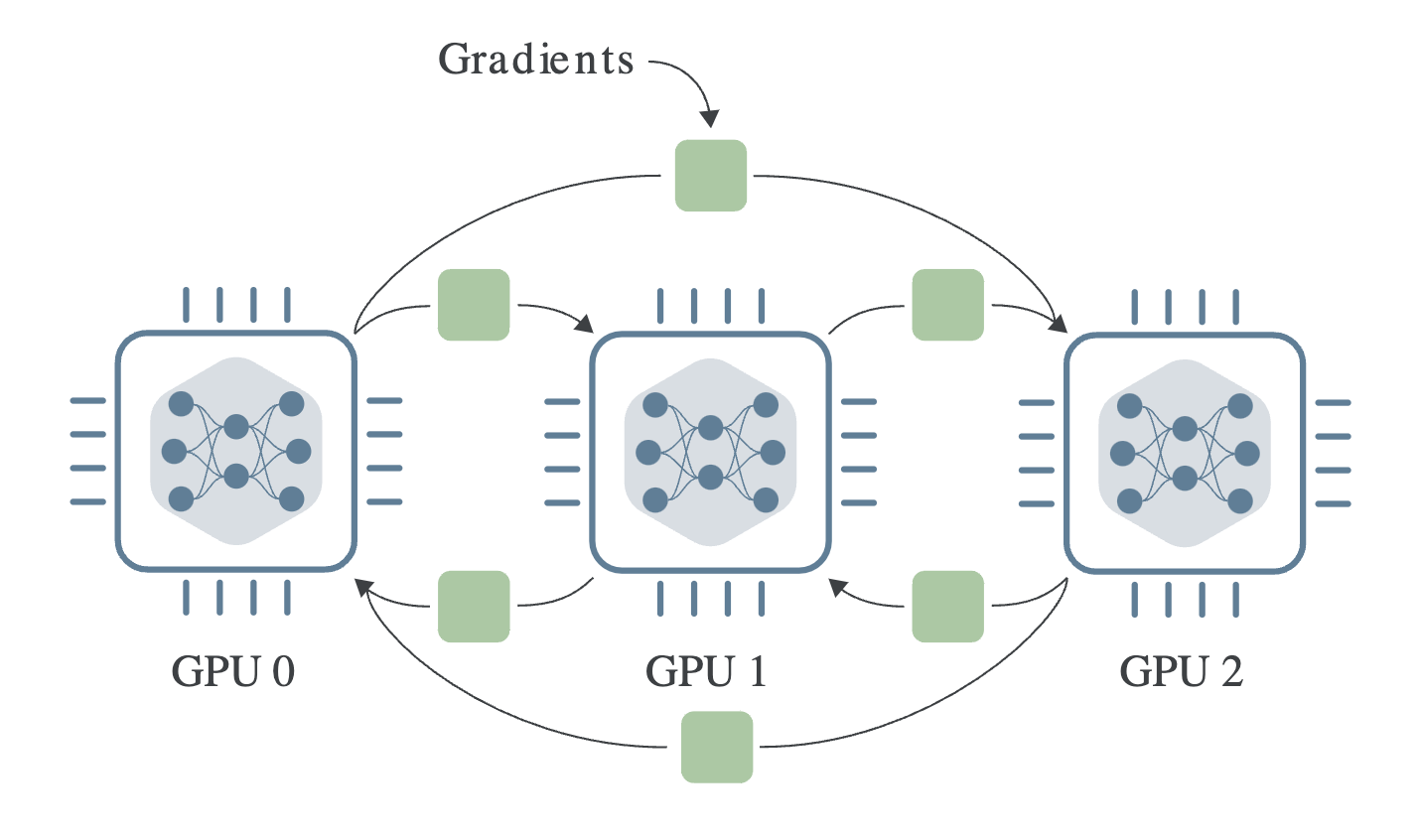}}
\label{DP}
\caption{Data Parallelism Gradient Computation \cite{tinkerd}}
\end{figure}

Distributed training is when multiple GPUs are used to train a single model. The three common types are Data Parallelism, Model Parallelism, and Pipeline Parallelism \cite{tinkerd}. Data parallelism (DP) is commonly used when models fit within device memory, replicating model parameters across devices and distributing mini-batches among processes. Each process handles a subset of data, performs forward and backward propagation, and updates model parameters using averaged gradients \cite{tinkerd}. The average of the gradients is used to update the model weights on each device to ensure that all devices have the same set of training weights at the beginning of the next training step as shown in Figure~\ref{DP}. This exchange of gradients between devices is performed with an algorithm called AllReduce, executing a reduction operation on data distributed across multiple devices \cite{tinkerd}. When models exceed device memory, model parallelism (MP) and pipeline parallelism (PP) are employed. PP horizontally partitions the model across devices and uses micro-batching to manage pipeline bubbles \cite{zero}. However, PP introduces challenges, such as complexities in implementing tied weights and batch normalization, large batch size requirements affecting convergence, and memory inefficiency \cite{zero}. To combat memory challenges, Zero Redundancy Optimizer is one input parameter in DeepSpeed that eliminates memory redundancies by partitioning three model states (optimizer states, gradients, parameters) across data-parallel processes rather than replicating them \cite{zero-inf}. There is also an extension to ZeRO called ZeRO-Infinity, where it takes advantage of GPU, CPU, and NVMe memory to allow models to scale on limited resources as we become limited by the GPU memory wall \cite{zero-inf}. However, due to resource and time constraints, our work will focus on data parallelism without ZeRO.

There is a recent survey by Duan et al. \cite{rw1} that investigates recent advancements in distributed systems for training LLMs like GPT and LLaMA, which demand extensive GPU clusters and significant computational resources. The paper reviews innovations in AI accelerators, networking, storage, and scheduling, alongside strategies for parallelism and optimizations in computation, communication, and memory usage. While it is a comprehensive survey on training language models, it shifts the focus on system reliability for long-duration training by exploring alternative computing approaches such as optical computing, which focuses more on hardware optimizations. Additional work in distributed training by Dash et al. \cite{rw2} examines efficient strategies for training trillion-parameter LLMs using Frontier, the first exascale supercomputer for open science. It evaluates model and data parallel techniques, including tensor parallelism, pipeline parallelism, and sharded data parallelism, focusing on their impact on memory, communication latency, and computational efficiency. The study identifies optimized strategies through empirical analysis and hyperparameter tuning, achieving high throughput and strong scaling efficiencies (up to 89\%) for large-scale models on thousands of GPUs. While these findings are significant for language models, there has not been work done in the Vision Transformers space to evaluate scalability and efficiency.

\section{Solution Design}
We use DeepSpeed \cite{b3} along with NCCL \cite{nccl} and MPI (OpenMPI implementation) to run on the remote clusters. Mpirun initializes a distributed environment in which each process (one per GPU) can communicate between nodes. MPI provides the rank and world size (total number of processes) that are crucial for inter-nodes communications. On the other hand, Deepspeed's \textit{init\_distributed()} initializes the training environment within each process. When called, it sets up Pytorch distributed backend (NCCL in our case) to allow processes to communicate data across GPUs, leveraging MPI configuration. In summary, MPI handles the distribution and launching of processes across nodes while DeepSpeed, in conjunction with Pytorch’s torch.distributed, manages the GPU-to-GPU communication and synchronization. torch.distributed.barrier() ensures that all processes synchronize at the end of each epoch, preventing any processes from moving to the next epoch before others are complete.

\begin{figure}[htbp]
\centerline{\includegraphics[width=\linewidth]{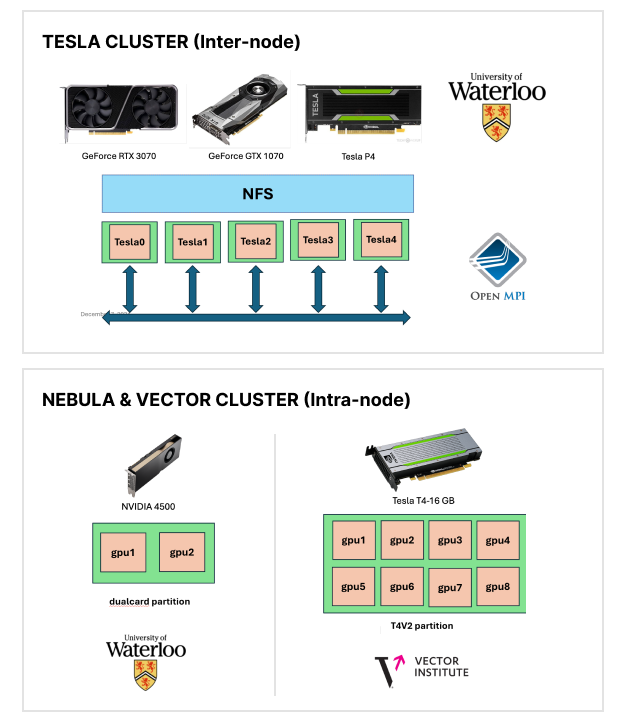}}
\caption{Remote Cluster Setup}
\label{ClusterConfigs}
\end{figure}

We train on three remote clusters to cover both intra- and inter-node training: Nebula, Tesla, and Vector. The configurations of the three clusters along with their corresponding GPUs are shown in Figure~\ref{ClusterConfigs}. Nodes are indicated in green and GPUs (within a node) are indicated in orange. On the Nebula and Vector machines, we use virtual environments from Anaconda. On the Tesla machine, we install non-pip virtual environment first, and manually install pip later. All the installing steps along with the code can be found in our Github repository \href{https://github.com/trinhgiahuy/Scalable_ViT_DT/tree/main}{trinhgiahuy/Scalable\_ViT\_DT}. It is important to note that due to resource limitations, the Tesla cluster does not have homogeneous GPUs for all five nodes, where machines 1, 2 and 4 have the same GPUs (RTX 3070) while machines 0 and 3 have weaker GPUs (GTX 1070 and Tesla P4).

We begin with intra-node training to test the functionality of the pipeline before moving onto inter-node training. We first implement the pipeline and test on the Nebula Cluster System from the ECE Linux computing facility (\textit{ece-nebula07.eng.uwaterloo.ca}). The pipeline has 2 modes: intra-node and inter-node training. On the Nebula cluster, we can use \textbf{\textit{midcard}} and \textbf{\textit{dualcard}} partitions. The \textbf{\textit{midcard}} partition comprises 1 node with 1 GPU while \textbf{\textit{dualcard}} partition comprises 2 nodes with 2 GPUs. Due to resource constraints, only 1 node is available in the \textbf{\textit{dualcard}} partition. We run the pipeline on 2 GPUs on the same node (share-memory system) first and compare it with single-node training. After the pipeline is functioning on Nebua, we deploy our implementation to \href{https://ece.uwaterloo.ca/Nexus/arbeau/clients/}{ECE GPU Ubuntu Servers}, where we use 6 machines: \textbf{eceubuntu0} as a controlling node and \textbf{ecetesla[0-4]} as compute nodes.
% [include something about scaling to multi-node training on Tesla?]
We use the Vision Transformer architecture \href{https://pytorch.org/vision/main/models/generated/torchvision.models.vit_b_16.html}{ViT\_b\_16} \cite{b1} and train on the datasets summarized in Table~\ref{dataset}. Finally, we deploy our work on Vaughan cluster from Vector Institute which has 54 nodes with 32 cores, 152GB memory, and 8 Tesla T4 (16GB) GPUs.

\begin{table}[htbp]
    \centering
    \caption{Datasets for evaluation}
    \begin{tabular}{|c|c|c|c|}
        \hline
        Dataset & No. of Classes & No. of Images & Resolution \\ \hline
        CIFAR-10 \cite{cifar} & 10 & 60,000 & 32x32 \\ \hline
        CIFAR-100 \cite{cifar} & 100 & 60,000 & 32x32 \\ \hline
        ImageNet-100 \cite{imagenet}* & 100 & 100,000 & 224x224 \\ \hline
    \end{tabular}
    \label{dataset}
    \footnotesize{*Due to time limitations, we could not train successfully on ImageNet, but the intention is to choose a dataset with higher resolution.}
\end{table}

\section{Evaluation}
\subsection{Evaluation Methodology}
The evaluation process involves conducting a series of experiments and scaling the number of GPUs to observe the trends in training time, communication overhead, and accuracy. We modify software parameters such as training batch size to monitor how the training performance changes across inter-node and intra-node setups. The key question is to assess whether DeepSpeed's data parallelism can effectively handle the computational demands of ViTs while maintaining efficiency and scalability.

Throughout the test, we fix the DeepSpeed configuration \ref{appendix:deepspeed_config} across all experiments and change the \textit{train\_batch\_size} and \textit{micro\_batch\_per\_gpu} accordingly. We demonstrate strong and weak scaling by modifying this configuration file in relation to the data set size. Strong scaling is achieved by fixing the workload by using the entire dataset for increasing number of GPUs. Weak scaling is achieved by modifying the partition of the dataset proportional to the number of GPUs so each GPU would receive equal workload. For example, 1 GPU uses 10\% of the dataset while 8 GPUs use 80\% of the dataset (each GPU will only compute on 10\% of the dataset as the number of GPUs scales). Since the training time should not vary between epochs for data parallelism, the model is trained for 5 epochs for all experiments and averaged when plotting the results (time in seconds) to dampen the effects of any outliers. 

% \subsection{Amdahl's Scaling Law}
% For inter-node distributed training, specifically for demonstrating strong scaling, we perform an analytic study by calculating the theoretical training time using Amdahl’s Law [CITE], which models the impact of sequential and parallel components on speedup:

% \begin{equation}
% \label{eq: Amdahl}
% T_N = T_1 \times \left( S + \frac{1 - S}{N} \right)
% \end{equation}
% where:
% \begin{itemize}
%     \item \( T_N \) is the training time on \( N \) GPUs,
%     \item \( T_1 \) is the training time on a single GPU,
%     \item \( S \) is the sequential fraction of the workload,
%     \item \( N \) is the number of GPUs.
% \end{itemize}

% \textbf{Speedup} is the ratio of the single GPU training time to the multi-GPU training time:

% \begin{equation}
% \label{eq: speedup}    
% \text{Speedup}_N = \frac{T_1}{T_N}
% \end{equation}

\subsection{Initial Scaling Results for Inter-node on Tesla}
% minor detail but the headings don't seem correct for some of these (weak scaling should be strong scaling). double check
\begin{figure}[htbp]
\centerline{\includegraphics[width=\linewidth]{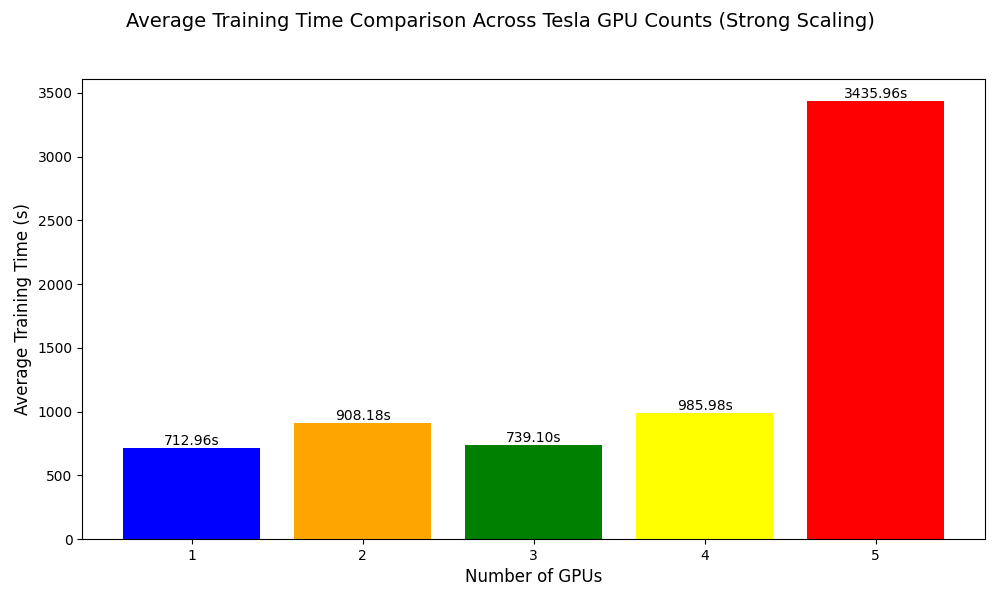}}
\caption{Tesla Strong Scaling}
\label{Tesla_Strong}
\end{figure}

\begin{figure}[H]
\centerline{\includegraphics[width=\linewidth]{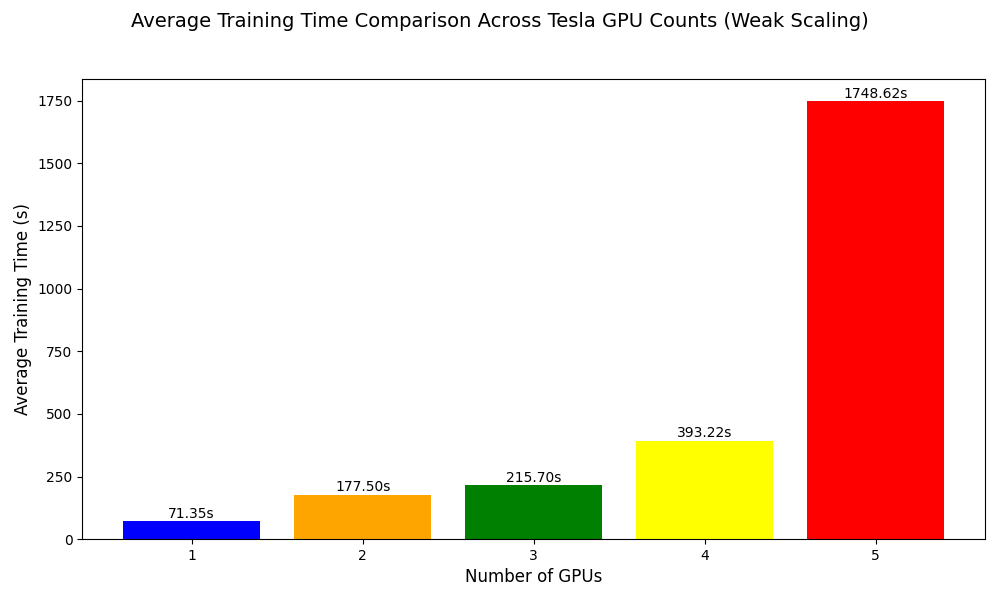}}
\caption{Tesla Weak Scaling}
\label{Tesla_Weak}
\end{figure}

Our first experiments uses Tesla machines for inter-node training, as shown in Figure~\ref{Tesla_Strong} and Figure~\ref{Tesla_Weak}. The results deviates from ideal strong or weak scaling, largely due to communication overhead between GPUs in the cluster. Two Tesla machines with weaker GPUs are limited to small batch sizes (e.g., 16), causing high synchronization costs due to frequent gradient averaging. Adding the fourth and fifth GPUs further increases training time because these weaker GPUs introduces computational bottlenecks, forcing other GPUs to wait during synchronization. This highlights the importance of GPU homogeneity for efficient scaling.

\subsection{Evaluating Communication Overhead on Nebula}
\begin{figure}[htbp]
\centerline{\includegraphics[width=\linewidth]{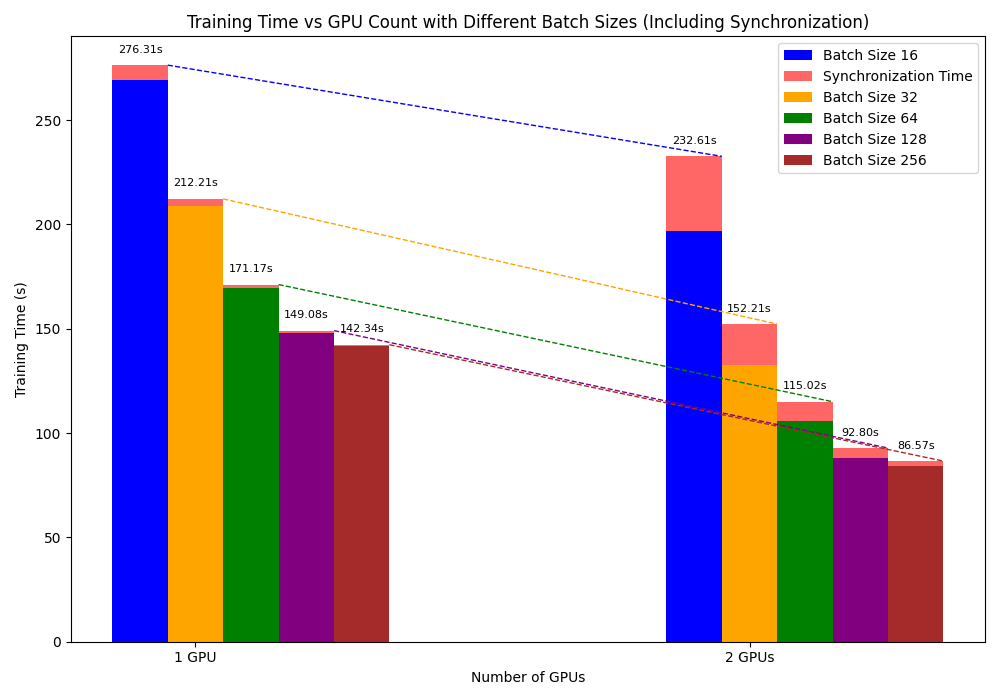}}
\caption{Nebula Strong Scaling vs Batch Sizes}
\label{Nebula_Strong}
\end{figure}

To examine the impact of batch size on synchronization costs, we switched to Nebula machines with more powerful GPUs capable of handling larger batch sizes. As shown in Figure~\ref{Nebula_Strong}, synchronization costs (highlighted in red) decreases significantly with larger batch sizes, especially in the two-GPU setup. Small batch sizes (e.g., 16) results in disproportionately high synchronization costs, leading to poor scaling. However, improvements plateau when the batch size increases from 128 to 256. One plausible explanation is that the GPU resources are already fully utilized, and larger batch sizes introduced a new bottleneck: the overhead of loading large batches from CPU to GPU memory \cite{b2}.

\begin{figure}[htbp]
\centerline{\includegraphics[width=\linewidth]{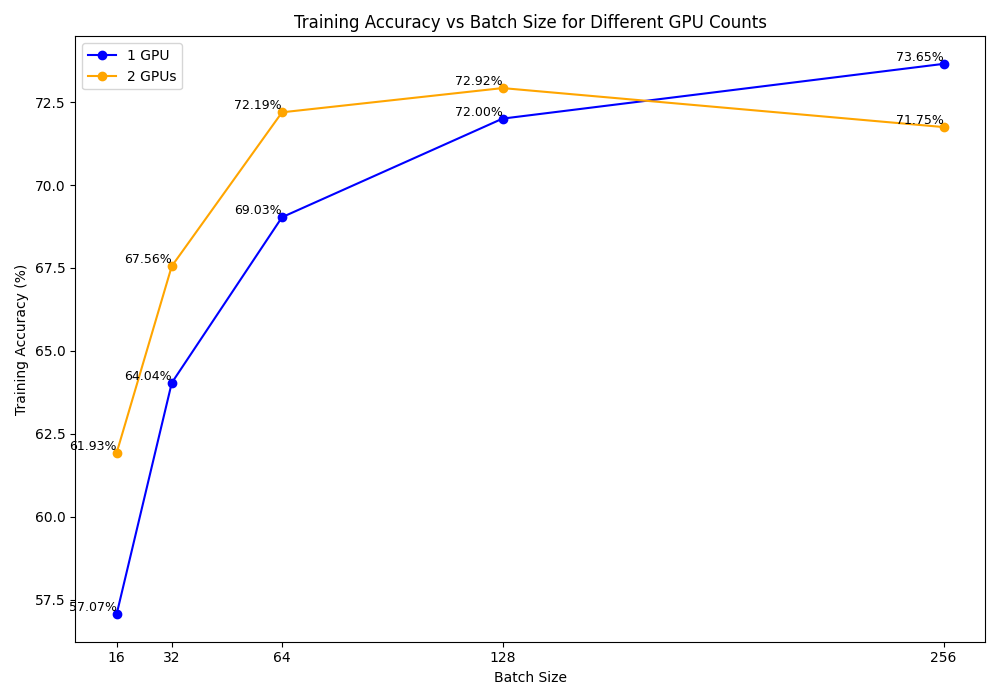}}
\caption{Nebula Accuracy vs Batch Sizes}
\label{Nebula_Accurancy}
\end{figure}

Figure~\ref{Nebula_Accurancy} illustrates how accuracy changes with batch size. Initially, the accuracy improves with increasing batch size but declines when the batch size became too large, possibly due to overfitting. This suggests an optimal balance exists between batch size and model performance.

Our findings suggest that a batch size of 64 or 128 offers a good trade-off between synchronization costs and memory usage for training Vision Transformer models with DeepSpeed. Additionally, gradient accumulation could be a promising parameter for GPUs with memory limitations, enabling effective larger batch sizes without frequent gradient averaging.

\subsection{Scaling Results for Intra-node on Vector}

\begin{figure}[H]
\centerline{\includegraphics[width=\linewidth]{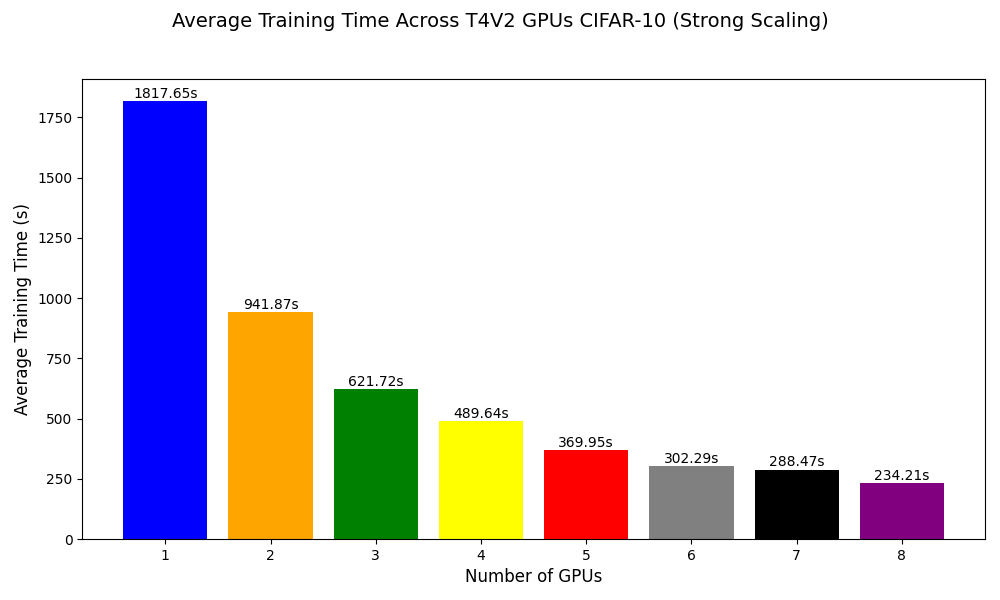}}
\caption{CIFAR-10 Strong Scaling (Batch size 64)}
\label{C10_Strong}
\end{figure}

\begin{figure}[H]
\centerline{\includegraphics[width=\linewidth]{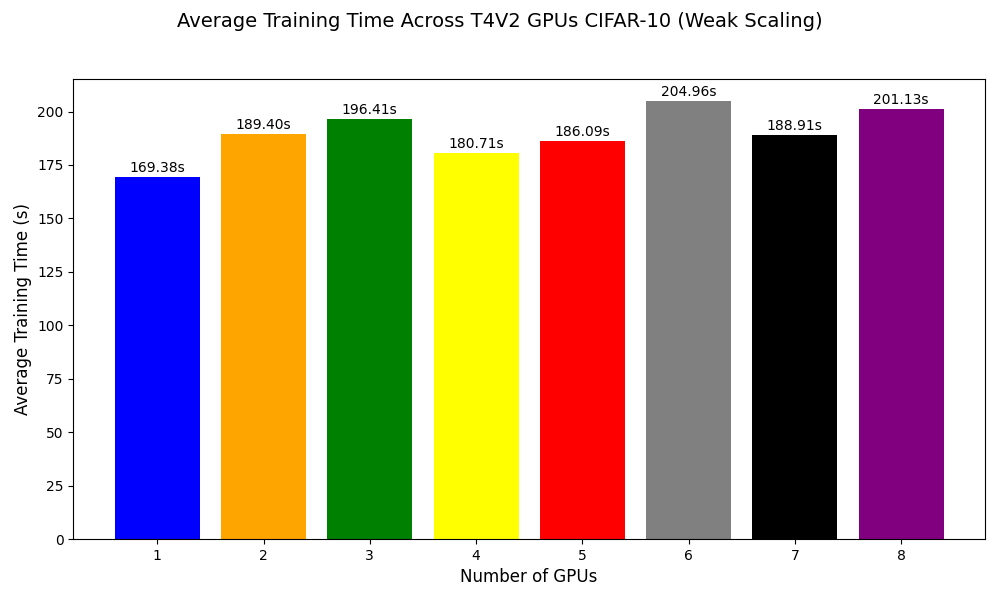}}
\caption{CIFAR-10 Weak Scaling (Batch size 64)}
\label{C10_Weak}
\end{figure}

Figure~\ref{C10_Strong} and Figure~\ref{C10_Weak} shows the trend for strong and weak scaling for CIFAR-10 dataset on T4V2 GPUs on the Vector cluster. As expected, strong scaling shows a consistent decrease in training time as more resources/GPUs are allocated for training. The scaling is best from 1 GPU to 2 GPUs, showing a reduction by almost half of the time. Weak scaling results are also as expected as we observe the times to remain constant for increasing GPUs.

\begin{figure}[htbp]
\centerline{\includegraphics[width=\linewidth]{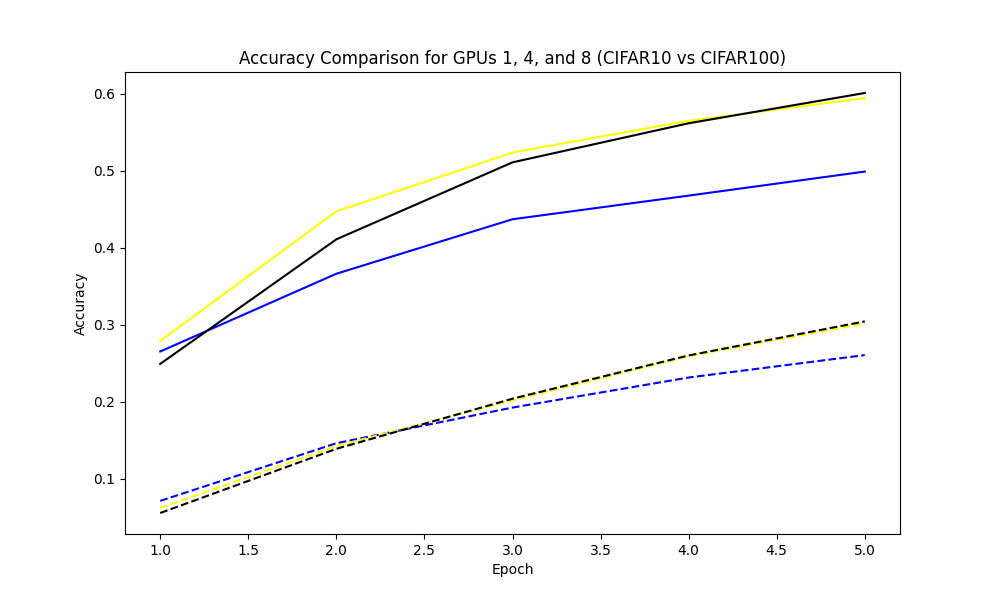}}
\caption{CIFAR-10 and CIFAR-100 Accuracies}
\label{C10/100_Acc}
\end{figure}

\begin{figure}[htbp]
\centerline{\includegraphics[width=\linewidth]{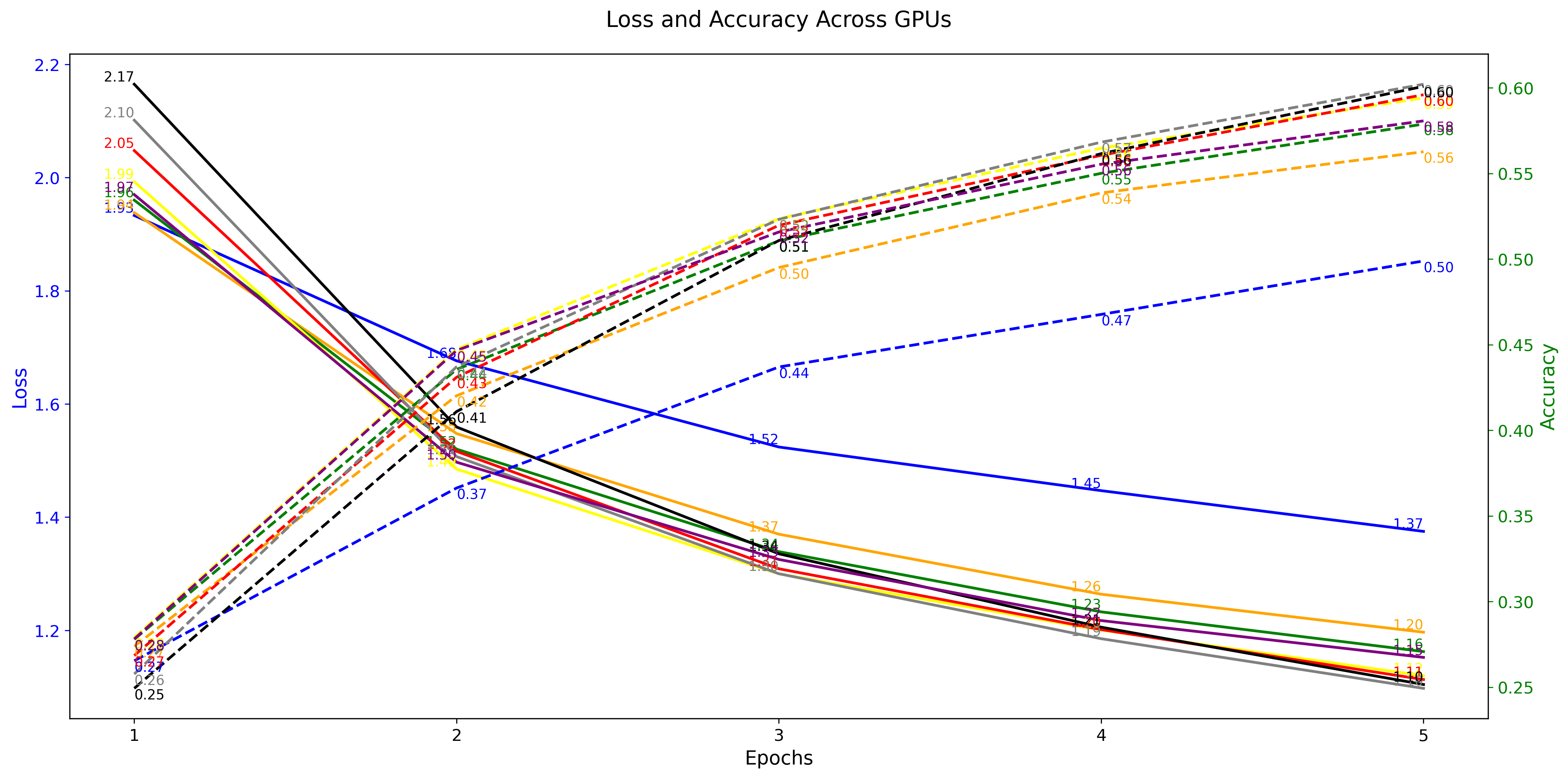}}
\caption{Loss and Accuracy of Strong Scaling (Batch size 64)}
\label{Loss/Acc}
\end{figure}

Similar results for the CIFAR-100 dataset are shown for strong and weak scaling as reported in Figure~\ref{C100_Strong} and Figure~\ref{C100_Weak} in the Appendix. This is expected since CIFAR-10 and CIFAR-100 have the same number of samples and same image resolutions. However, when comparing the accuracies of the two, it is expected that the CIFAR-100 performs worse in comparison since it has more classes and less data to train for each class. The accuracy comparisons between the two datasets are shown in Figure~\ref{C10/100_Acc}, where it is also observed that 4 GPUs and 8 GPUs has a better performance than 1 GPU for later epochs. This does not necessarily mean that more GPUs result in a higher accuracy, but it is an observation from our results. Figure~\ref{Loss/Acc} shows successful distributed training as loss decreases and accuracy increases for each epoch.

\begin{figure}[H]
\centerline{\includegraphics[width=\linewidth]{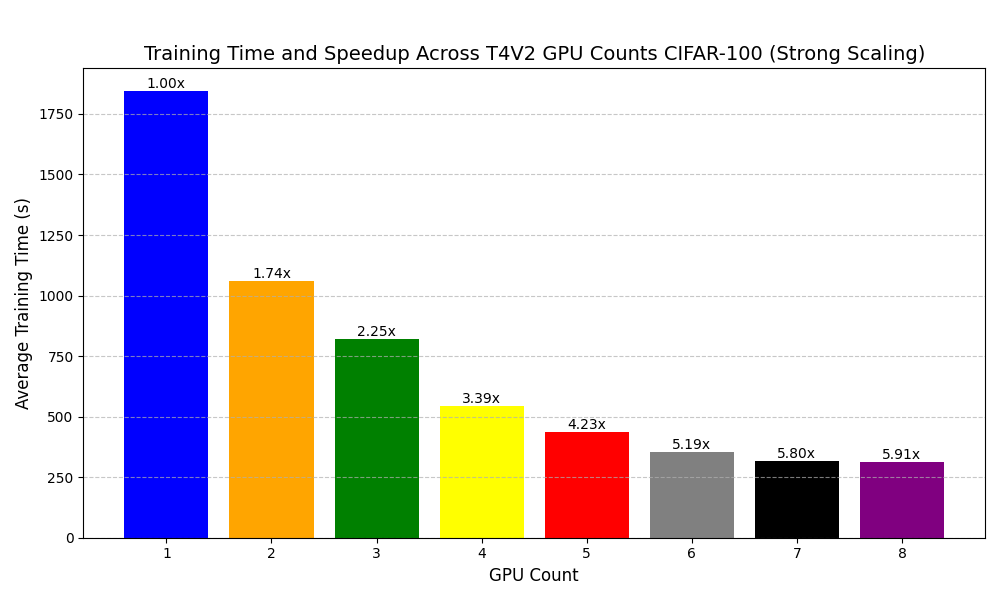}}
\caption{CIFAR-10 Strong Scaling Speedup (Batch size 16)}
\label{C100_SU_b16}
\end{figure}

\begin{figure}[H]
\centerline{\includegraphics[width=\linewidth]{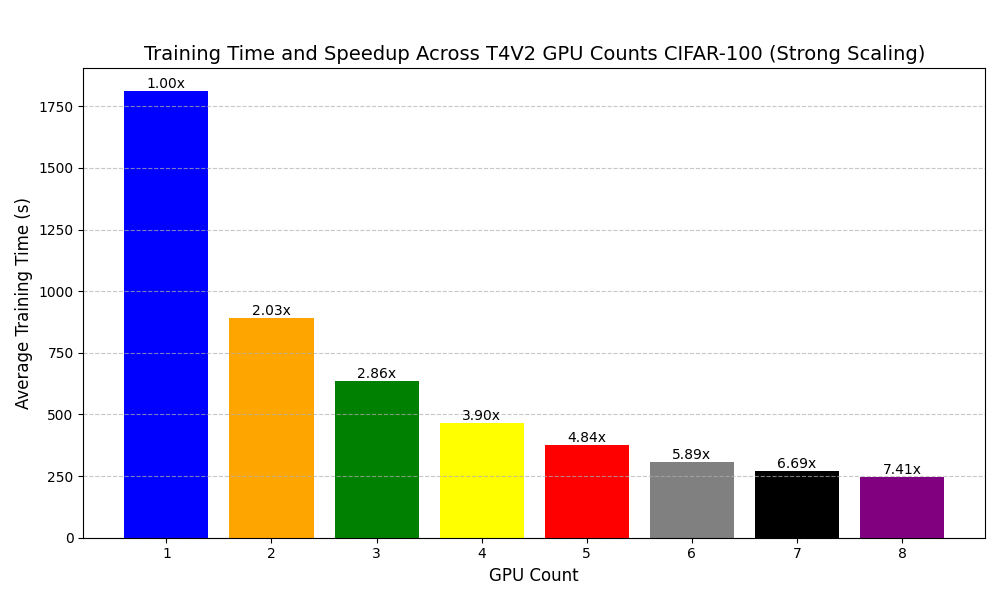}}
\caption{CIFAR-10 Strong Scaling Speedup (Batch size 64)}
\label{C100_SU_b64}
\end{figure}

From the results from Nebula cluster, we found that the optimal batch size that produces the best results for distributed training for ViTs is 128. However, due to memory constraints in the Vector cluster, only batch size of 64 can be successfully obtained. All the results presented thus far are using batch size 64, but we also experimented with batch size 16 to compare the speedup for the CIFAR-100 dataset. Figure~\ref{C100_SU_b16} and Figure~\ref{C100_SU_b64} shows the speedup of batch sizes 16 and 64, respectively. As expected, the speedup ratio is generally better for the larger batch size.

\subsection{Scaling Results for Inter-node on Vector}
We also conduct Multi-Node Single-GPU experiments on the Vector cluster. Specifically, we use only 1 GPU per node and scale the total GPU count from 1 to 32 nodes. The training workload on CIFAR-100 is fixed, focusing on strong scaling.

\begin{figure}[H]
\centerline{\includegraphics[width=\linewidth]{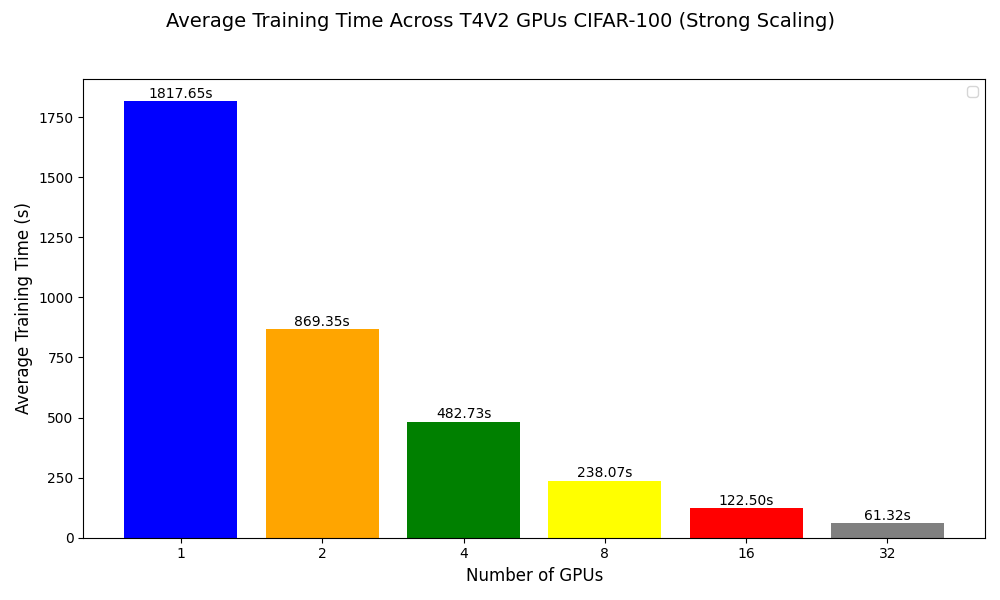}}
\caption{Multi-node Single GPU Strong Scaling Result On CIFAR-100 (Batch size 64)}
\label{inter-node-C100_Strong}
\end{figure}

Figure \ref{inter-node-C100_Strong} demonstrates the strong scaling of Multi-node Single GPU up to 32 nodes.

% \begin{table}[h]
% \caption{Multi-node Single GPU Strong Scaling Efficiency Result.}
% \small % Changes the font size to small
% \setlength{\tabcolsep}{5pt} % Adjusts the spacing between columns
% \begin{tabular}{l|c|c|c}
% \toprule
%      % & Number of MPI Processes & Number of OpenMP threads  & Domain Decomposition\\
%   GPU(s)        & Actual Time (s) & Theoratical Time(s) & Efficiency (\%) \\
% \midrule
% 1  &  1817.654000   &   1817.654000 & 100.000000  \\
% 2  & 869.348750 & 999.709700 & 104.541129 \\
% 4 & 482.727500 & 590.737550 & 94.134579 \\
% 8 & 238.071500 & 386.251475 & 95.436350 \\
% 16 & 61.315125 & 232.886919 & 92.638949 \\
% % \bottomrule
% \label{tab:efficient}
% \end{tabular}
% \end{table}

% Interestingly, Figure \ref{inter-node-C100_Strong} shows that the training time result consistently outperforms the theoretical predictions.This gap narrows as GPU count increases, suggesting that DeepSpeed mitigates communication overhead effectively, even in a distributed setup.
% In particular, the actual training time achieves a speedup of 29.64x on 32 GPUs compared to a single GPU with efficiency remaining above 92\%. 

\begin{figure}[H]
\centerline{\includegraphics[width=\linewidth]{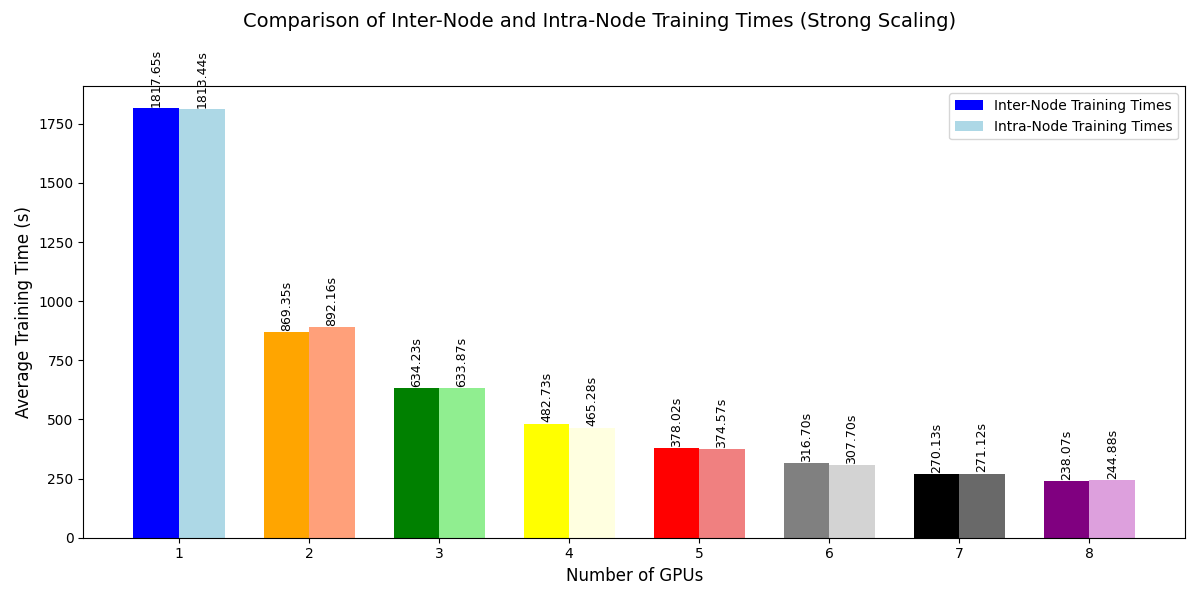}}
\caption{Multi-node Single-GPU vs Single-Node Multi-GPU Strong Scaling Result On CIFAR-100 (Batch size 64)}
\label{inter-intra-C100_Strong}
\end{figure}

Finally, we compare the strong scaling results between Multi-node Single-GPU (in bold colors) and Single-Node Multi-GPU (in light colors) using batch size of 64. Figure~\ref{inter-intra-C100_Strong} shows that there are no significant differences between using GPU intra-node and inter-node. 

\section{Conclusion and Future Work}
This study explored the use of DeepSpeed, a distributed training framework, to improve the scalability and performance of Vision Transformers (ViTs) for image classification tasks. Our evaluations, conducted across various GPU setups (e.g., Nebula, Tesla, and Vector clusters) and datasets such as CIFAR-10 and CIFAR-100, revealed key insights into training speed, communication overhead, and demonstrated strong and weak scaling when using distributed data parallelism for both inter- and intra-node training. While the Tesla inter-node training was unsuccessful due to non-homogeneous GPUs, the Vector inter-node training showed good scaling rseults, scaling to 32 nodes. Through intra-node training on Nebula and Vector clusters, we demonstrated the effect of software parameters like batch size on training efficiency. We determined that batch sizes of 64 or 128 optimally minimize synchronization costs, achieving better speedup while effectively utilizing GPU memory. We also showed that there is more communication overhead when scaling to multiple GPUs, which is important to keep in mind when managing the demands of large-scale models. By optimizing software parameters and addressing hardware limitations, we can achieve efficient scaling and optimal training accuracy in ViTs. This study is only the beginning to investigating distributed training for ViTs and we hope to continue our work with some future improvements.

% this is like immediate work for benchmarking
Immediate future works include conducting further experiments to understand the limitations of DeepSpeed on Vision Transformers. For instance, ViTs often have large intermediate activations due to high resolution images, which might strain memory more than token-based LLMs. To extend our work, we can evaluate each ZeRO stage to measure memory savings and overhead, as well as test performance with different optimizers such as SGD and LAMB \cite{lamb}. We could also benchmark the performance against other distributed training frameworks such as Megatron-LM or HuggingFace Accelerate when applied to ViTs.

% this is future works that is more novel
Finally, our work can be extended to Vision Language Model or scientific imaginary research, especially the scalability of processing very long sequence of images. DeepSpeed-Ulysses, which uses model sharding parallelism and ZeRO-3 optimization, enables highly efficient LLM training with long sequence lengths \cite{ulysses}. The authors propose sequence parallelism as a solution to partition the input sequence along the sequence length dimension using all-to-all communication for attention computation \cite{ulysses}. We will adapt this to our project by tokenizing image patches in the Vision Transformer model. Instead of token-based sequences, we would partition along the "image patches dimension" and combine with recent advanced research of vision models such as Long-Sequence-Segmentation \cite{hsvit}, SparseViT \cite{sparsevit}, etc for further sequence parallelism scaling. We would evaluate our work on medium and high resolution or multiple-channel images dataset (e.g., "fastMRI" \cite{fastmri} in medical imaging, "CoSTAR" \cite{costar} for robotics, "GTDB" \cite{gtdb} for genomics, etc).

% not sure if we want an acknowledgement xd

% \clearpage
\appendices
\section{Vector Cluster Results}
\begin{figure}[H]
\centerline{\includegraphics[width=\linewidth]{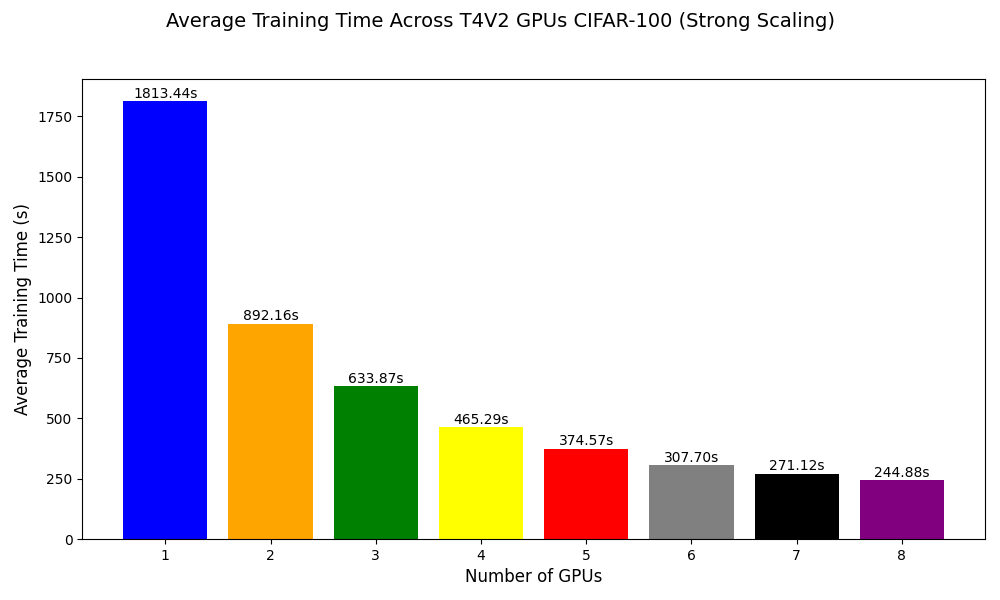}}
\caption{CIFAR-100 Strong Scaling (Batch size 64)}
\label{C100_Strong}
\end{figure}

\begin{figure}[H]
\centerline{\includegraphics[width=\linewidth]{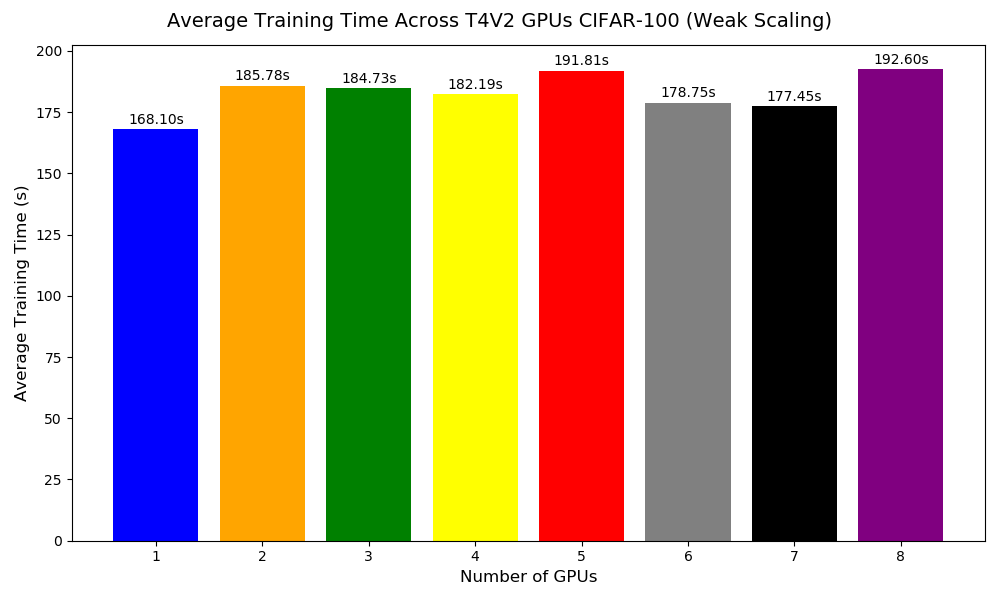}}
\caption{CIFAR-100 Weak Scaling (Batch size 64)}
\label{C100_Weak}
\end{figure}

\section{DeepSpeed Configuration Code}
\label{appendix:deepspeed_config}
\lstset{language=JSON}
\begin{lstlisting}
{
  "train_batch_size": 32,
  "gradient_accumulation_steps": 1,
  "micro_batch_per_gpu": 16,
  "fp16": {
    "enabled": false
  },
  "zero_optimization": {
    "stage": 0,
    "offload_optimizer": {
        "device": "none"
    },
    "offload_param": {
        "device": "none"
    }
},
"wall_clock_breakdown": true,
"prescale_gradients": false,
"pipeline": {
    "pipe_partitioned": false
},
"pin_memory": true
}
\end{lstlisting}
\end{document}